\documentclass[conference]{IEEEtran}
\IEEEoverridecommandlockouts

\usepackage{cite}
\usepackage{amsmath,amssymb,amsfonts}
\usepackage{algorithmic}
\usepackage{graphicx}
\usepackage{textcomp}
\usepackage{xcolor}
\usepackage{subfigure}
\usepackage{hyperref}
\def\BibTeX{{\rm B\kern-.05em{\sc i\kern-.025em b}\kern-.08em
    T\kern-.1667em\lower.7ex\hbox{E}\kern-.125emX}}
\begin{document}

\title{Language Models as Efficient Reward Function Searchers for Custom-Environment Multi-Objective Reinforcement
}

\author{
\IEEEauthorblockN{
Guanwen Xie\IEEEauthorrefmark{1},
Jingzehua Xu\IEEEauthorrefmark{1},
Yiyuan Yang\IEEEauthorrefmark{2},
Yimian Ding\IEEEauthorrefmark{1}$^{,+}$,
Shuai Zhang\IEEEauthorrefmark{3}$^{,+}$
}
\IEEEauthorblockA{\IEEEauthorrefmark{1}Tsinghua Shenzhen International Graduate School, Tsinghua University, China}
\IEEEauthorblockA{\IEEEauthorrefmark{2}Department of Computer Science, University of Oxford, United Kingdom}
\IEEEauthorblockA{\IEEEauthorrefmark{3}Department of Data Science,
New Jersey Institute of Technology, USA}
Email: sz457@njit.edu
\thanks{ $^{+}$ These authors contributed equally to this work. 

$^{1}$$\,$The full-text prompts, examples of LLM-generated answers, and source code are available at https://360zmem.github.io/LLMRsearcher/ .} 
}

\maketitle

\begin{abstract}
    Achieving the effective design and improvement of reward functions in reinforcement learning (RL) tasks with complex custom environments and multiple requirements presents considerable challenges. In this paper, we propose ERFSL, an efficient reward function searcher using LLMs, which enables LLMs to be effective white-box searchers and highlights their advanced semantic understanding capabilities. Specifically, we generate reward components for each numerically explicit user requirement and employ a reward critic to identify the correct code form. Then, LLMs assign weights to the reward components to balance their values and iteratively adjust the weights without ambiguity and redundant adjustments by flexibly adopting directional mutation and crossover strategies, similar to genetic algorithms, based on the context provided by the training log analyzer. We applied the framework to a customized data collection RL task without direct human feedback or reward examples (zero-shot learning). The reward critic successfully corrects the reward code with only one feedback instance for each requirement, effectively preventing unrectifiable errors. The initialization of weights enables the acquisition of different reward functions within the Pareto solution set without the need for weight search. Even in cases where a weight is 500 times off, on average, only 5.2 iterations are needed to meet user requirements. The ERFSL also works well with most prompts utilizing GPT-4o mini, as we decompose the weight searching process to reduce the requirement for numerical and long-context understanding capabilities$^{1}$.
    \end{abstract}
    \section{Introduction}
    \label{sec:intro}

    Reinforcement learning (RL) methods are being increasingly utilized for intricate, multi-objective tasks. Nonetheless, as the variety and quantity of requirements and optimization goals grow, the design of reward functions becomes more complex, necessitating significant effort to adjust the structure and coefficients of each reward component. This complexity is further compounded by the fact that researchers' needs often fluctuate with different scenarios and over time, and can sometimes be ambiguous \cite{4}, thereby posing a considerable challenge to achieving optimal performance.

    Large language models (LLMs) are trained on extensive text data \cite{13}, enabling them to demonstrate strong problem-solving and content generation abilities when given well-written prompts, even without prior domain knowledge. The application of LLMs in creating functional code has shown remarkable performance across various tasks, such as dexterous agents control \cite{2,8,17} and Minecraft playing \cite{5,7}, highlighting their potential in zero-shot scenarios with limited self-evolution iterations. However, this improvement process depends on trial-and-error exploration. When there is a single, clear objective (e.g., success rate), the search space for designing functions and tuning parameters is confined, allowing for gradual optimization through iterations. However, for complex reward functions, where reward components and their weights are determined simultaneously, issues such as incorrect code and imbalanced weights may arise, which are difficult to resolve solely through training feedback. To address these challenges, some approaches decompose complex tasks into several sub-tasks or skills, while providing clear task feedback accordingly \cite{21,22,23}. 
    
    Furthermore, multi-objective RL tasks face the challenge of balancing the weights and values in the reward function. A related topic is LLM-driven white-box numerical optimization \cite{13,3,10}, such as hyperparameter optimization (HPO) \cite{3,19}. This approach involves abstracting all explicitly defined coefficients and the function code itself into parameters, allowing LLMs to perform comprehensive analyses and improvements on well-defined machine learning tasks. This paradigm aligns with the observation that LLMs are particularly adept at summarizing and heuristically generating code within specific and clear task contexts, yet they are less proficient in addressing black-box optimization problems \cite{1}. 
    
    \begin{figure*}[!t]
        \centering
        \includegraphics[width=0.902\linewidth]{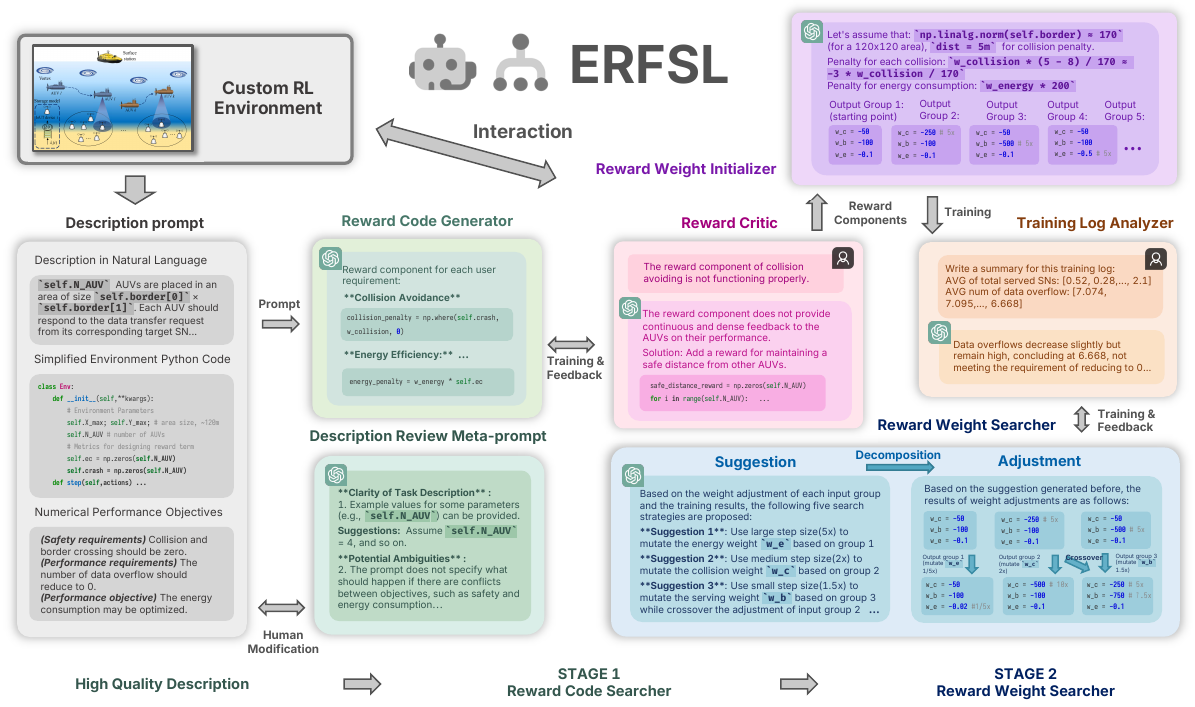}
        \caption{The main architecture and prompt examples of the proposed ERFSL framework.}
        \label{fig_1}
        \end{figure*}
    
    These considerations underscore the necessity of providing clear task descriptions while rationally exploring the extensive search space. We propose ERFSL, an efficient reward function searcher via LLMs for custom-environment multi-objective RL, which utilizes task decomposition and white-box search to fully leverage LLMs' semantic understanding capabilities. Unlike previous work, we separate the reward-generating process into reward code design and reward weight searching, breaking down the RL task into explicit numerical goals to eliminate ambiguity in training feedback. We implement a clear feedback and self-evolution search paradigm, using a reward critic to rectify reward components for each user requirement, subsequently optimizing the weight coefficients of these components. We apply ERFSL to the reward function design for a customized data collection task from our previous research \cite{14}, achieving user requirements with minimal iterations and weight searches. Notably, aside from the task description, ERFSL operates without direct human feedback (easy for writing automatic scripts). However, it can also be effectively combined with human knowledge. For instance, by using weight searchers to optimize the weights and values of human-designed reward functions. The main architecture and important prompt examples are illustrated in Fig. 1.
    
    \label{sec:format}

    \section{Methodology}
    

    \subsection{Environment Description and Reward Code Generaton} 
    \label{subsec:cp}
    Task description is a common part of most subsequent prompts, comprising text descriptions, environment code, or APIs, which include variables and functions essential for designing the reward function, as well as user requirements. We decompose the user requirements into numerical-clear performance demands (e.g., obstacle avoidance to achieve zero collision). Ambiguous task descriptions may  hinder LLMs' ability to generate correct reward functions. To facilitate human modifications, we design a meta-prompt to provide suggestions for enhancing the quality of the description. This meta-prompt enables LLMs to identify potential issues within the prompt, such as unclear structural organization and a lack of necessary information or explanations.
    
    The code generation process can directly borrow the existing LLM-aided reward function design frameworks. However, instead of creating a single reward function for the entire output, we generate individual reward components tailored to each specific user requirement. 
    
    The initial code produced by LLMs is likely to be incorrect due to the absence of prior knowledge about custom environments and the complexity of contexts. Therefore, we test each component separately and correct errors using a LLM-based reward critic. The reward critic follows a step-by-step guide, namely first lists possible reasons for code failure, then reviews the environment code and the requirement, and finally outputs the correct function code. This process allows LLMs to clearly analyze errors within the reward function, thereby avoiding the ambiguity of overall feedback. In addition, if the variables and functions are insufficient to write new components, LLMs are allowed to fabricate relative variables and prompt the user to complete them, which effectively overcome the negative effects caused by incomplete environment descriptions.

    \subsection{Reward Weight Search}
    
    Multi-objective reinforcement learning necessitates not only the proper form of reward components but also their reasonable scaling. We utilize large language models (LLMs) as effective weight searchers under explicit task contexts. First, we leverage a reward weight initializer to provide an effective search starting point, which requires LLMs to pre-calculate approximate values of the components and adjust the weights to ensure these components' values are on the same scale. This approach prevents the initial weight groups from deviating significantly from the optimal solutions. Afterward, the reward weight searcher offers new weight solutions based on the training results. Existing methods utilize Python list-style training logs to present these results. However, detailed logs make the prompt excessively lengthy and complex, significantly hindering textual and numerical understanding, especially for small-scale LLMs \cite{20,24}. Consequently, we employ a training log analyzer to produce a concise textual summary of the task.
    
    To facilitate directional adjustment and minimize the impact of errors in LLM decision-making, the reward weight searcher processes K=5 input weight groups and generates K=5 output groups. The reward weight initializer designates the first weight group as the starting point for the search, with the other four sets subsequently adjusting certain weights of the first set. Subsequently, the reward weight searcher provides suggestions for weight adjustment based on training results summarized by the training log analyzer. To prevent ambiguity and potential redundancies in weight adjustments across multiple input sets, the searcher specifies starting points for mutation (adjustment) on each weight from a certain input group, allowing for directional modifications like maintaining, increasing/decreasing, or fine-tuning weights. For multiple weight adjustments, we conduct crossovers between the modified weight groups to integrate changes made in mutated input groups relative to the search start point. While borrowing the terminologies from the genetic algorithm and inspired by it, our process ensures directed mutation based on input groups and selective crossovers to enhance search efficiency.
    
    The complex search strategies arising from multiple weight groups lead to lengthy prompts. Similarly, by separating the processes of generating adjustment suggestions and new output weight groups, we ensure the necessary understanding for the former and precise execution for the latter.
    
    \section{Experiments}
    \label{sec:exp}
    
    \subsection{Task Description and Parameters} 
    To evaluate the proposed ERFSL framework, we select the customized data collection task from our previous work \cite{14}, which utilizes the RL algorithm to control multiple custom-simulated mobile agents (AMAs) for data collection. We task LLMs with designing reward functions encompassing safety requirements (collision and border crossing avoidance), performance requirements (timely serving of the target SN to minimize data overflow), and performance objectives (reduce energy consumption), without providing any reward examples (namely zero-shot), as shown in the task description in Fig. 1. We refer to the original paper for the system models and simulation parameters. For determinacy, TD3 \cite{18} is used as the RL algorithm instead of MAISAC, as in the original paper.
    
    For experiments, we utilize \verb|gpt-4o-2024-08-06| (denoted as \textbf{GPT-4o}) as the default LLM due to its improved performance and reasonable API pricing. We also conduct experiments using the cost-efficient smaller version of OpenAI's LLMs, \verb|gpt-4o-mini-2024-07-18| (denoted as \textbf{GPT-4om}). The LLM parameters are set to \verb|temperature=0.5| and \verb|Top P=1|. Similar to Eureka \cite{2}, we designed a prompt that takes the reward functions and numerical values as a whole, and this baseline is called as \textbf{EUREKA-S}. Meanwhile, we set a baseline called \textbf{EUREKA-M}, which processes one or more reward functions along with their training logs summarized by the training log analyzer as inputs during the reward function revision stage, and generates K=5 outputs simultaneously. This approach remains consistent with the reward weight searcher, in contrast to generating a single reward function repeatedly in an i.i.d. manner as in Eureka.
    
    
    
    \subsection{Main Results and Case Studies} 
    \label{subsec:mr}
    
    \textbf{Reward Critic can generate correct code stably and quickly.} The initially generated reward functions may not be entirely correct, as they may only contain sparse reward terms, miscalculate distances between AMAs and the boundary, and contain symbolic errors, among other issues. To address these problems, the reward critic generates feedback and revised code for each component. We utilize a variation of the training log analyzer, which outputs \verb|[YES]| and \verb|[NO]| to indicate whether a specific component fulfills the requirement. We find that the reward critic can effectively detect various errors and then rewrite reward components based on the description and variables of the environment class. For each component, only one iteration is needed to obtain the correct feedback results.
    
    \begin{figure}[!t]
        \centering
        \includegraphics[width=0.958\linewidth]{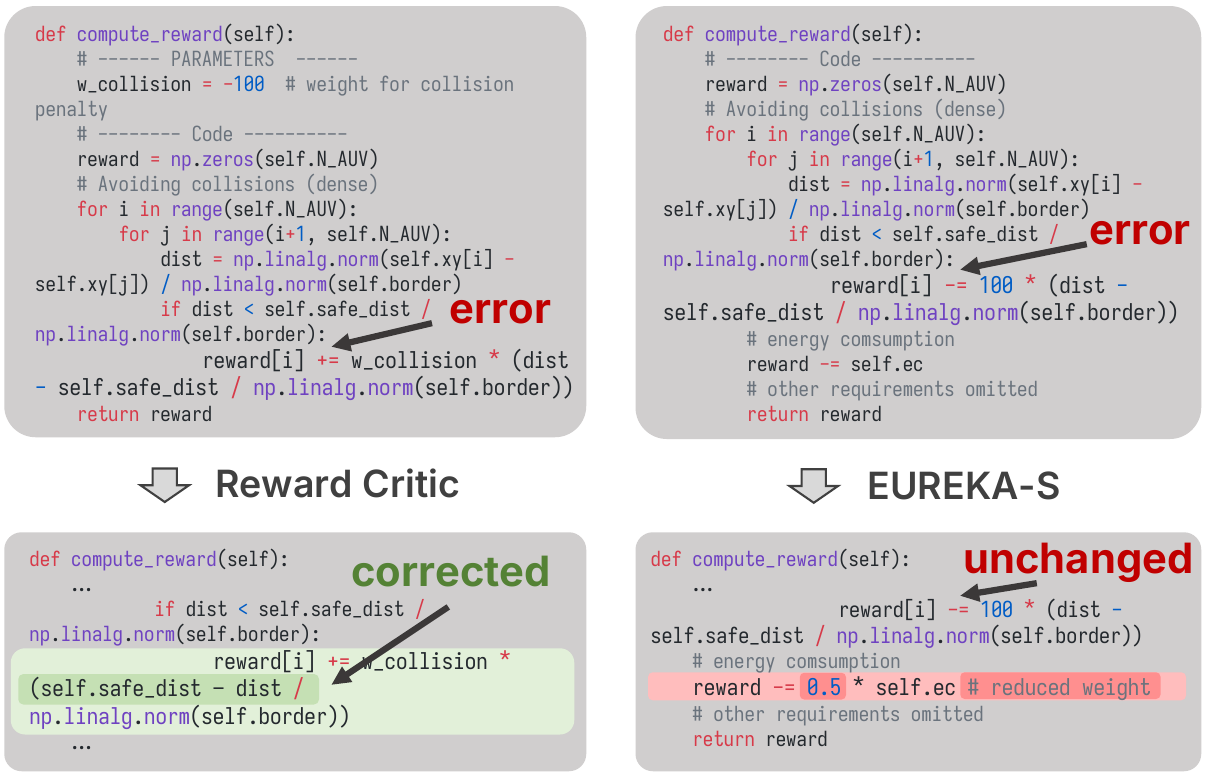}
        \vspace{-3mm}
        \caption{When the collision penalty term is reversed to a reward term, the reward critic corrects the error, but directly feeding back the entire reward function only results in weight modifications.}
        \label{fig_2}
        \vspace{-3mm}
        \end{figure}

    To investigate the advantages of the reward critic, we manually introduce an elusive error into the reward function by reversing the penalty term for collision into a reward term, namely reversing the symbol \verb|reward -= ...| to \verb|reward += ...|. Example outputs of the reward critic and EUREKA-S are shown in Fig. 2. The reward critic can identify the error, but EUREKA-S fails to do so, only modifying reward weights, even when prompted with "This code contains errors." We also attempted to use the sparse term-only reward function for EUREKA-S to modify, but unless explicitly prompted to generate a dense term, it only continues with weight modification. This suggests that the reward design process based on training feedback may be ineffective without explicit human input.

    \begin{table}[t]
            \centering
            \caption{The number of iterations of searching weights under different experiment settings. The experiment are performed 5 times, and the mean values and standard deviations are reported.}
            \label{table:masking_performance}
            \vspace{3mm}
            \begin{tabular}{|l|c|c|c|} 
            \hline
            Setting & \textbf{GPT-4o} & \textbf{GPT-4o w/o TLA} & \textbf{GPT-4om}  \\
            \hline
            \textbf{ERFSL}    & \multicolumn{3}{|c|}{\textbf{0.40±0.49}}  \\
            \hline
            \textbf{ERFSL w/o balance}  & 1.20±0.75 & 1.60±1.02 & 1.80±1.17 \\ 
            \hline
            \textbf{ERFSL 500x off}  & 5.20±1.46 & 6.40±1.86 & 8.60±1.74  \\ 
            \hline
            \end{tabular}
            \vspace{-1mm}
            \end{table}

            \begin{figure}[!t]
            \centering
            \subfigure[Solutions generated from the reward weight initializer.]{\includegraphics[width=0.49\linewidth]{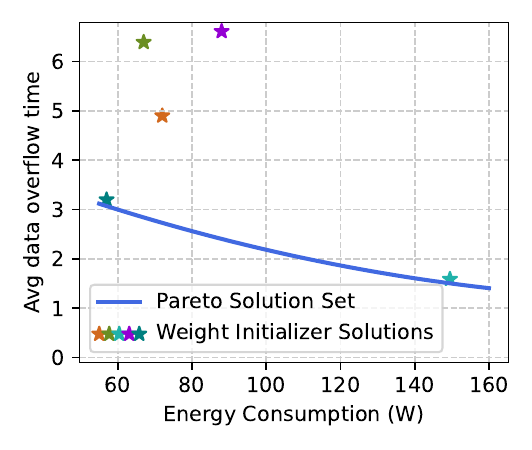}\label{fig: sub_figure1}} 
            \subfigure[Change of maximum value of w\_service/w\_ec under different settings during iteration.]{\includegraphics[width=0.49\linewidth]{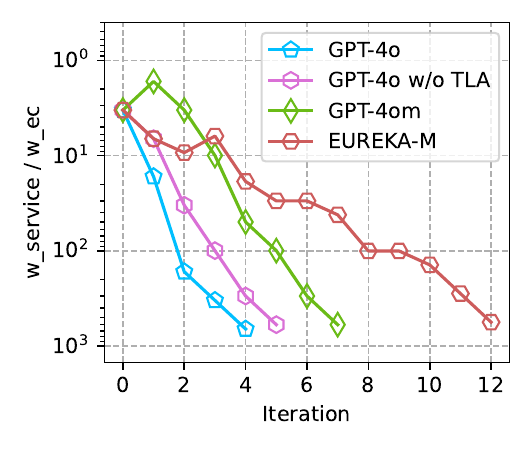}\label{fig: sub_figure2}}
            \caption{Figures of reward weight searching. (a) Solutions generated from the reward weight initializer. (b) Change of step sizes under different settings during iteration.}
            \label{fig: figure}
        \end{figure} 
    
    \textbf{Reward weight initialization and search.} We obtain initial groups of weights from the reward weight initializer, and the energy consumption and average data overflow times (lower is better) of these solutions are illustrated in Fig. 3(a). Although three groups of generated weights do not meet user requirements, two groups (emphasizing timely response to target SNs and energy consumption, respectively) successfully achieve Pareto solutions, indicating that no further search is required, or only a more refined search may be necessary for a specific point on the Pareto set. Then, we employ the reward weight searcher to iteratively adjust the weight coefficients. We also remove the value balance process of the weight initializer (i.e., eliminating the example values from the environment description and removing prompts requiring LLMs to balance reward values, denoted as \textbf{ERFSL w/o balance}) to perform comparative experiments. We also perform ablation experiments by removing the training log analyzer (i.e., inputting original training logs directly, denoted as \textbf{GPT-4o w/o TLA}). Additionally, we task LLMs to search from a weight group generated by the reward weight initializer, but with the weight of the energy consumption penalty term increased by a factor of 500 (denoted as \textbf{500x off}), to understand the search details. Table 1 displays the number of iterations required to meet user demand, while Fig. 3(b) shows the maximum value of \verb|w_service/w_ec| in five weight groups during training, with the best performance among the five repetitions. When the weight initialization does not consider the balance between reward components, the ratio between weights compared to RWI-initialized settings could differ by a factor of 1 to 50, leading to a significant increase in the deviation of generated weights. Nevertheless, since the distance is not substantial, only 0-3 iterations are necessary to find a feasible solution.

    In the 500x off experiment, \textbf{GPT-4o} (GPT-4o with TLA) quickly recognizes excessive energy consumption optimization. It initially uses a 5x search step size, then increases it to expedite the process, and finally reduces it once data overflows decrease, resulting in an average of 5.2 iterations. When the training log analyzer is removed, the choice of step size in complex situations is less flexible compared to \textbf{GPT-4o}. We also utilize EUREKA-M as a pure weight searcher for the weight searching process. EUREKA-M adopts small, random step sizes and increases weights in a highly random manner rather than decreasing the penalty of energy consumption, necessitating a substantial number of iterations.

    \textbf{The difference between utilizing GPT-4o and GPT-4om.} Intuitively, GPT-4om and open-source LLMs exhibit weaker overall reasoning abilities compared to GPT-4o. Their numerical analysis and mathematical capabilities are also comparatively limited, which leads to poor performance in designing reward functions for agentic control \cite{16}. This limitation results in drawbacks for utilizing GPT-4om in ERFSL, such as an improperly functioning reward function initializer. When responding to content-generating prompts (such as code design and training log analysis) exceeding 5k tokens, the quality of GPT-4om is inferior to that of GPT-4o. Nevertheless, GPT-4om still performs adequately. Similar to GPT-4o, the reward critic requires only one feedback iteration to correct the code per requirement. Furthermore, due to the decomposition of the reward weight searching process, GPT-4o effectively proposes suggestions and completes the process with performance surpassing that of EUREKA-M. However, the step sizes provided by GPT-4om lack flexibility, requiring additional search iterations.

    \section{Conclusion and Discussion} 
    \label{sec:con}
    
    In this paper, we propose ERFSL, which decomposes a multi-objective task into clear user requirements, enabling LLMs to function as zero-shot searchers that receive clear feedback and effectively generate reward functions. LLMs are tasked to generate reward components, which are subsequently corrected by the reward critic to prevent potential errors. Leveraging the enhanced numerical calculation capabilities of the latest GPT-4o, we initialize weights that balance the value of reward components, allowing us to obtain feasible Pareto solutions without the need for extensive searching. Furthermore, drawing from non-numerical contexts provided by the training log analyzer, GPT-4o can flexibly adopt different search strategies, including directional mutation and crossover, thereby accelerating the search process. In addition, with the exception of the reward weight initializer, the most prompts exhibit acceptable performance in GPT-4om. Future work may focus on developing clearer and more automated task descriptions and on verifying the LLM-aided reward design process across a broader range of tasks.

    \bibliographystyle{IEEEtran}

\begin{thebibliography}{10}
\providecommand{\url}[1]{#1}
\csname url@samestyle\endcsname
\providecommand{\newblock}{\relax}
\providecommand{\bibinfo}[2]{#2}
\providecommand{\BIBentrySTDinterwordspacing}{\spaceskip=0pt\relax}
\providecommand{\BIBentryALTinterwordstretchfactor}{4}
\providecommand{\BIBentryALTinterwordspacing}{\spaceskip=\fontdimen2\font plus
\BIBentryALTinterwordstretchfactor\fontdimen3\font minus \fontdimen4\font\relax}
\providecommand{\BIBforeignlanguage}[2]{{%
\expandafter\ifx\csname l@#1\endcsname\relax
\typeout{** WARNING: IEEEtran.bst: No hyphenation pattern has been}%
\typeout{** loaded for the language `#1'. Using the pattern for}%
\typeout{** the default language instead.}%
\else
\language=\csname l@#1\endcsname
\fi
#2}}
\providecommand{\BIBdecl}{\relax}
\BIBdecl

\bibitem{4}
C.~F. Hayes, R.~R{\u{a}}dulescu, E.~Bargiacchi, J.~K{\"a}llstr{\"o}m, M.~Macfarlane, M.~Reymond, T.~Verstraeten, L.~M. Zintgraf, R.~Dazeley, F.~Heintz \emph{et~al.}, ``A practical guide to multi-objective reinforcement learning and planning,'' \emph{Autonomous Agents and Multi-Agent Systems}, vol.~36, no.~1, p.~26, 2022.

\bibitem{13}
S.~Liu, C.~Chen, X.~Qu, K.~Tang, and Y.-S. Ong, ``Large language models as evolutionary optimizers,'' \emph{arXiv preprint arXiv:2310.19046}, 2023.

\bibitem{2}
Y.~J. Ma, W.~Liang, G.~Wang, D.-A. Huang, O.~Bastani, D.~Jayaraman, Y.~Zhu, L.~Fan, and A.~Anandkumar, ``Eureka: Human-level reward design via coding large language models,'' in \emph{The Twelfth International Conference on Learning Representations}, 2024.

\bibitem{8}
Y.~Zeng, Y.~Mu, and L.~Shao, ``Learning reward for robot skills using large language models via self-alignment,'' \emph{arXiv preprint arXiv:2405.07162}, 2024.

\bibitem{17}
W.~Yu, N.~Gileadi, C.~Fu, S.~Kirmani, K.-H. Lee, M.~G. Arenas, H.-T.~L. Chiang, T.~Erez, L.~Hasenclever, J.~Humplik \emph{et~al.}, ``Language to rewards for agentic skill synthesis,'' \emph{arXiv preprint arXiv:2306.08647}, 2023.

\bibitem{5}
Y.~Wu, S.~Y. Min, S.~Prabhumoye, Y.~Bisk, R.~R. Salakhutdinov, A.~Azaria, T.~M. Mitchell, and Y.~Li, ``Spring: Studying papers and reasoning to play games,'' \emph{Advances in Neural Information Processing Systems}, vol.~36, 2024.

\bibitem{7}
H.~Li, X.~Yang, Z.~Wang, X.~Zhu, J.~Zhou, Y.~Qiao, X.~Wang, H.~Li, L.~Lu, and J.~Dai, ``Auto mc-reward: Automated dense reward design with large language models for minecraft,'' in \emph{IEEE/CVF Conference on Computer Vision and Pattern Recognition}, 2024.

\bibitem{21}
Z.~Mandi, S.~Jain, and S.~Song, ``Roco: Dialectic multi-robot collaboration with large language models,'' in \emph{2024 IEEE International Conference on Robotics and Automation (ICRA)}.\hskip 1em plus 0.5em minus 0.4em\relax IEEE, 2024, pp. 286--299.

\bibitem{22}
E.~Triantafyllidis, F.~Christianos, and Z.~Li, ``Intrinsic language-guided exploration for complex long-horizon robotic manipulation tasks,'' in \emph{2024 IEEE International Conference on Robotics and Automation (ICRA)}.\hskip 1em plus 0.5em minus 0.4em\relax IEEE, 2024, pp. 7493--7500.

\bibitem{23}
S.~Rho, L.~Smith, T.~Li, S.~Levine, X.~B. Peng, and S.~Ha, ``Language guided skill discovery,'' \emph{arXiv preprint arXiv:2406.06615}, 2024.

\bibitem{3}
M.~Zhang, N.~Desai, J.~Bae, J.~Lorraine, and J.~Ba, ``Using large language models for hyperparameter optimization,'' in \emph{NeurIPS 2023 Foundation Models for Decision Making Workshop}, 2023.

\bibitem{10}
Z.~Ma, H.~Guo, J.~Chen, G.~Peng, Z.~Cao, Y.~Ma, and Y.~jiao Gong, ``Llamoco: Instruction tuning of large language models for optimization code generation,'' \emph{arXiv preprint arXiv:2403.01131}, 2024.

\bibitem{19}
C.~Wang, X.~Liu, and A.~H. Awadallah, ``Cost-effective hyperparameter optimization for large language model generation inference,'' in \emph{International Conference on Automated Machine Learning}.\hskip 1em plus 0.5em minus 0.4em\relax PMLR, 2023, pp. 21--1.

\bibitem{1}
B.~Huang, X.~Wu, Y.~Zhou, J.~Wu, L.~Feng, R.~Cheng, and K.~C. Tan, ``Exploring the true potential: Evaluating the black-box optimization capability of large language models,'' \emph{arXiv preprint arXiv:2404.06290}, 2024.

\bibitem{14}
Z.~Zhang, J.~Xu, G.~Xie, J.~Wang, Z.~Han, and Y.~Ren, ``Environment- and energy-aware ama-assisted data collection for the information updating networks,'' \emph{IEEE Internet of Things Journal}, vol.~11, no.~15, pp. 26\,406--26\,418, 2024.

\bibitem{20}
T.~Li, G.~Zhang, Q.~D. Do, X.~Yue, and W.~Chen, ``Long-context llms struggle with long in-context learning,'' \emph{arXiv preprint arXiv:2404.02060}, 2024.

\bibitem{24}
W.~Shen, C.~Li, H.~Chen, M.~Yan, X.~Quan, H.~Chen, J.~Zhang, and F.~Huang, ``Small llms are weak tool learners: A multi-llm agent,'' \emph{arXiv preprint arXiv:2401.07324}, 2024.

\bibitem{18}
S.~Fujimoto, H.~Hoof, and D.~Meger, ``Addressing function approximation error in actor-critic methods,'' in \emph{International conference on machine learning}.\hskip 1em plus 0.5em minus 0.4em\relax PMLR, 2018, pp. 1587--1596.

\bibitem{16}
H.~Chen, F.~Jiao, X.~Li, C.~Qin, M.~Ravaut, R.~Zhao, C.~Xiong, and S.~Joty, ``Chatgpt's one-year anniversary: are open-source large language models catching up?'' \emph{arXiv preprint arXiv:2311.16989}, 2023.

\bibitem{6}
T.~Xie, S.~Zhao, C.~H. Wu, Y.~Liu, Q.~Luo, V.~Zhong, Y.~Yang, and T.~Yu, ``Text2reward: Reward shaping with language models for reinforcement learning,'' in \emph{The Twelfth International Conference on Learning Representations}, 2024.


\end{thebibliography}


\end{document}